\documentclass[sigconf,anonymous=false,nonacm=true]{acmart}
\usepackage{tikz}
\usetikzlibrary{bayesnet}
\usetikzlibrary{arrows}
\usetikzlibrary{backgrounds}
\usetikzlibrary{calc}
\usepackage{graphicx}
\usepackage{subcaption}
\usepackage{algorithm}
\usepackage{algorithmicx}
\usepackage[noend]{algpseudocode}
\usepackage{caption}
\usepackage{amsfonts}
\usepackage{amsmath}

\begin{document}

\title{Multi-modal multi-objective model-based genetic programming to find multiple diverse high-quality models}

\author{E.M.C. Sijben}
\affiliation{%
 \institution{Centrum Wiskunde \& Informatica}
 \city{Amsterdam}
 \country{the Netherlands}
}
\email{evi.sijben@cwi.nl}

\author{T. Alderliesten }
\affiliation{%
 \institution{Leiden University Medical Center}
 \city{Leiden}
 \country{the Netherlands}
}
\email{t.alderliesten@lumc.nl}

\author{P.A.N. Bosman }
\affiliation{%
 \institution{Centrum Wiskunde \& Informatica}
 \city{Amsterdam}
 \country{the Netherlands}
}
\email{peter.bosman@cwi.nl}

\begin{abstract}
Explainable artificial intelligence (XAI) is an important and rapidly expanding research topic. The goal of XAI is to gain trust in a machine learning (ML) model through clear insights into how the model arrives at its predictions. Genetic programming (GP) is often cited as being uniquely well-suited to contribute to XAI because of its capacity to learn (small) symbolic models that have the potential to be interpreted. Nevertheless, like many ML algorithms, GP typically results in a single best model. However, in practice, the best model in terms of training error may well not be the most suitable one as judged by a domain expert for various reasons, including overfitting, multiple different models existing that have similar accuracy, and unwanted errors on particular data points due to typical accuracy measures like mean squared error. Hence, to increase chances that domain experts deem a resulting model plausible, it becomes important to be able to explicitly search for multiple, diverse, high-quality models that trade-off different meanings of accuracy. In this paper, we achieve exactly this with a novel multi-modal multi-tree multi-objective GP approach that extends a modern model-based GP algorithm known as GP-GOMEA that is already effective at searching for small expressions.

\end{abstract}

\maketitle

\begin{figure}
    \centering
    \includegraphics[width=0.44\textwidth]{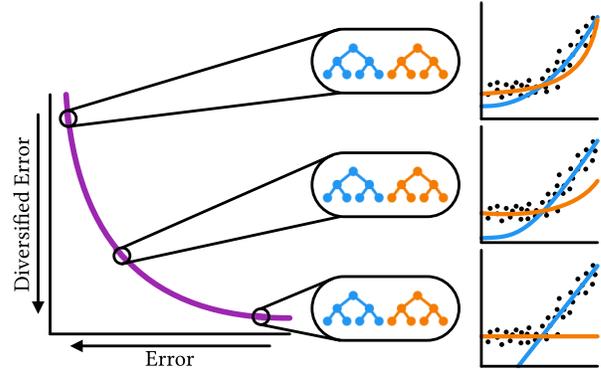}
    \vspace*{-3mm}
    \caption{Visualization of our approach.}
    \label{fig:eye_catcher}
    \vspace*{-3mm}
\end{figure}

\section{Introduction}
State-of-the-art machine learning models are often difficult to interpret.
This can be caused by models having many coefficients, many variables, and/or an intricate structure.
The popular field of eXplainable Artificial Intelligence (XAI) aims to either develop methods that enable humans to interpret the complex machine learning model and its predictions, or to make inherently interpretable models.
Models are unlikely to be inherently interpretable if they have many coefficients and/or variables.
Therefore, it makes sense to restrict the size of the model. However, doing so may negatively affect model performance.
Ideally, one would obtain both high performance and interpretability.

Genetic Programming (GP) is a learning algorithm that can generate expressions that are flexible in their structure, and in the operators or functions used in these expressions.
This enables GP to capture non-linear relations in a compact expression, offering a useful trade-off
between performance and size compared to other methods~\cite{la2021contemporary}.
This quality is why GP has been suggested to be useful for XAI~\cite{virgolin2021improving,evans2019s,ferreira2020applying}.

XAI approaches aim to give users insight into a model.
This allows users to make more informed use of the model in the real world~\cite{molnar2020interpretable,hleg2019ethics}.
However, if a model has (serious) shortcomings or other problems, although now the user may be informed about these potential problems, insight alone does not necessarily provide a solution, nor a direct way to change the model.
Of course, users can always choose not to use the model.
Ideally, however, users would be able to choose a model from a list of models that exhibit different qualities, to meet their specific preferences and prevent model rejection.
Unfortunately, this is typically not possible because most machine learning algorithms generate only one model.

In practice, due to the amount of data available (both in terms of records and variables) and/or the complexity of the prediction task, combined with a finite capacity of learnable models, it is unlikely that the optimal model (in terms of minimum training or validation error) is unique.
Even if it is, it might not be the best model according to the user. 
The user may have certain model requirements, or there could be conflicts between the model and expert knowledge about the domain.
Additionally, it might not be the best model simply because the data is limited and models with slightly worse performance in error on the data at hand are better if we had infinite data to train on.

Therefore, it would be very useful in practice, with the vision of a domain-expert end-user selecting the preferred model for the task at hand, to be able to search for multiple models that are all \emph{well-performing} but are also \emph{diverse} and potentially even describe different parts of the data better. 
Allowing the user to inspect such a set of different models may provide unique novel insights into the data and the process underlying the data, and support choosing a good model with additional expert knowledge in a powerful and sensible way.
Thereby user agency and control are increased.

In this paper, we propose exactly this: searching, in a novel way, for \emph{sets of multiple models} instead of a single model. Rather than using (adaptive) niching or fitness sharing, we use a multi-tree GP model.
This enables us to explicitly define diversity between models and perform multi-modal search in a potentially highly multi-modal search space by finding a fixed number of modes/niches.
Moreover, by defining the search in a Multi-Objective (MO) way, we can optimize for both diversity between models and model performance. Finally, by using particular notions of diversity, we can not only find models that are different, but even focus on different parts of the data in different ways, giving additional meaning and potentially practically useful dimensions to the models that will be presented to the user. 
We visualize our approach in \autoref{fig:eye_catcher}.
To the best of our knowledge, this is the first paper to propose searching for diverse multi-tree models with an MO approach.

Besides searching for a set of models, the ultimate goal is to create interpretable models.
While we do not directly focus on interpretability here, we do consider that smaller models are likely more interpretable than larger models. Under this assumption, it becomes interesting to look at GP approaches that are particularly well-suited to evolving small solutions. Results show that the GP variant of the Gene-pool Optimal Mixing Evolutionary Algorithm (GP-GOMEA) gives better results when searching for solutions of limited tree size than classic GP~\cite{virgolin2021improving}.
An MO version of GP-GOMEA does not yet exist, however.
Therefore, we implement an MO variant of GP-GOMEA by leveraging the best practices previously employed in the design of MO-GOMEA~\cite{luong2014multi}.

The contribution of this paper is threefold.
We 1) develop a novel approach for searching for sets of high-quality and diverse multi-tree GP models, 2) implement a version of GP-GOMEA with multi-trees and MO optimization, and 3) show the benefits of our approach by applying it to real-world data sets.
\vspace{-1em}

\section{Related work}
Diversity maintenance is a well-studied subject in GP that aims to improve diversity in the whole population to prevent premature convergence.
In ~\cite{galvan2019promoting} semantic diversity is promoted by using the semantic crowding distance. The semantic crowding distance adopts the same principles as the crowding distance except it does not look at the distance in objective space but in semantic space. Not only is the regular crowding distance replaced by the semantic one, the semantic crowding distance is also added as an objective.

In ~\cite{burks2015efficient}, an extension of the age-fitness Pareto optimization ~\cite{schmidt2011age}, an MO method that optimizes age and fitness, is proposed.
The idea behind this is to avoid that younger individuals, that have had less time to become fit, compete with older individuals, that have had more time to become fit. Thereby, the competition between individuals with a similar age is stimulated.
A metric for tree structure similarity (genetic marker density) is used instead of age to prevent converging to a specific structure. 

The FOCUS method~\cite{de2001reducing} performs an MO search with three objectives: fitness, size, and average squared overlapping tree distance to other members in the population. Individuals are stimulated to move away from the central peak in the fitness landscape if the central peak becomes too crowded.

In ~\cite{mckay2000fitness} fitness sharing for GP is introduced. The general idea is to reward solutions that are different, i.e., the reward for each prediction is divided by the number of individuals in the population that give the same prediction.
A distance function that reflects the structural dissimilarity of trees to extend the applicability of fitness sharing for tree-based methods is introduced in ~\cite{ekart2000metric}. 

Our work distinguishes itself from the above-mentioned works because we do not primarily aim to avoid premature convergence, but we aim to present the user with a diverse set of potentially interesting models.
Furthermore, the approach we present in this paper to realize this goal, is novel, for two reasons: 1) we maintain diversity within individuals rather than diversity across the population by using an MO search with multi-tree individuals, and 2) we incorporate a novel diversity objective in this approach that has particular relevance to machine learning.

\section{Multi-modal multi-objective model-based GP}
In this section, we present our approach to search for diverse high-quality models.
It has four key components: GP-GOMEA, multi-tree individuals, MO optimization, and a particular diversity objective.

We use GP-GOMEA~\cite{virgolin2021improving} because it is known to offer a good trade-off between performance and model size~\cite{la2021contemporary}.
We implement multi-tree individuals, which allows us to express diversity within individuals.
We choose to maintain diversity within individuals rather than maintaining diversity across the population because this allows us to more easily stimulate diversity and performance at the same time.
To optimize for both diversity and performance we implement an MO variant of GP-GOMEA, where we leverage best practices of MO-GOMEA~\cite{luong2014multi}, and introduce a diversity objective function, which we optimize together with an error objective.

\subsection{Gene-pool Optimal Mixing Evolutionary Algorithm (GOMEA)}
GOMEA is a model-based Evolutionary Algorithm (EA) that has been shown to be effective in many domains such as discrete optimization~\cite{thierens2011optimal,luong2014multi}, real-valued optimization~\cite{bouter2017exploiting}, and, most relevant here, GP~\cite{virgolin2021improving}. GOMEA differs from classic EAs in that it uses a linkage model that is meant to capture the interdependencies within the genotype. This information is then used during variation to prevent building blocks (or partial solutions) from being disrupted and to effectively mix these blocks to create better solutions.

In GOMEA, a fixed-length string is used as the genotype so that genes at a certain location in the string always represent the same variables in the problem. Linkage information is represented as a Family Of Subsets (FOS). The FOS contains subsets of genes (string indices) that are assumed to be linked.

If no linkage information is known a priori, it can be learned from the population during search. To this end, Mutual Information (MI) is often used to measure linkage among gene pairs and a so-called Linkage Tree (LT) is built to represent variable dependence relations in a hierarchical fashion. Computing joint MI for more than two genes is costly and requires large population sizes to be accurate. Therefore, the UPGMA algorithm~\cite{gronau2007optimal} is used to approximate linkage for larger groups of genes. UPGMA is a hierarchical clustering algorithm that merges subsets, starting from singleton sets, which in GOMEA contain the individual genes. For the similarity measure in UPGMA, the pairwise average MI is used. At each merge step, the newly constructed cluster is added to the FOS, ultimately resulting in the LT. Merging is usually stopped in GOMEA when only two clusters are left. The subset containing all genes is typically not added to the FOS because using this subset would result in entire solutions being cloned during variation.

In GOMEA, every generation, each individual in the population undergoes variation through Gene-pool Optimal Mixing (GOM). GOM uses the information in the FOS to replace linked genes at the same time. Suppose individual $\mathcal{P}_i$ undergoes GOM. First, $\mathcal{P}_i$ is cloned into offspring $\mathcal{O}_i$. Then, each of the subsets in the FOS is considered in a random order. For each FOS subset anew, a donor is randomly selected from the population. The values of the genes in $\mathcal{O}_i$ are replaced with those of the donor, but only at the positions specified by the FOS subset. If a replacement did not result in a worse fitness, the change is kept.

After processing all FOS elements, $\mathcal{O}_i$ is added to the offspring set and the next population member is considered. After processing the entire population, the population is replaced by the offspring.

\vspace*{-1mm}
\subsection{GP-GOMEA \& Multi-trees}
GP-GOMEA is a GP variant of GOMEA~\cite{virgolin2021improving}. In GP-GOMEA, individuals are trees that adhere to a template with fixed node positions. This enables mapping trees to strings in a fixed manner, i.e., nodes at the same position in different trees always map to the same string index. Consequently, learning an FOS and GOM variation can be straightforwardly used. However, a particular form of normalization is used on the estimated MI values to account for suprious linkage in the initial population that may occur in the case of GP trees because not every node in the GP tree is always used, nor can every node represent the same thing (leaves cannot represent functions). For more details, see~\cite{virgolin2021improving}.
 
\begin{figure}[!h]
  \centering
  	\begin{subfigure}[t]{0.8\linewidth}
		\centering
		\scalebox{0.5} {
		\begin{tikzpicture}[thick, every node/.style={scale=1.3}, every label/.style=draw]
    \node[latent, draw=none] (a) {$x_1$};%
    \node[latent,right= of a, label={[label distance=-0.5em, fill=white]30:4}] (b) {$x_2$}; %
    \coordinate (Middle1) at ($(a)!0.5!(b)$);
    \node[latent, draw=none, right=of b] (c) {$x_1$}; %
    \node[latent,right=of c, label={[label distance=-0.5em, fill=white]30:7}] (d) {$x_3$}; %
    \coordinate (Middle2) at ($(c)!0.5!(d)$);
    \node[latent, draw=none,above=of Middle1] (e) {$\times$}; %
    \node[latent,above=of Middle2, label={[label distance=-0.5em, fill=white]30:5}] (f) {$/$}; %
    \coordinate (Middle3) at ($(e)!0.5!(f)$);
    \node[latent,above=of Middle3, label={[label distance=-0.5em, fill=white]30:1}] (g) {$+$}; %
     
    % edges
    \draw (g) -- (e);
    \draw (g) -- (f);
    \draw (e) -- (a);
    \draw (e) -- (b);
    \draw (f) -- (c);
    \draw (f) -- (d);
     
    \node[latent, label={[label distance=-0.5em, fill=white]30:3}] (a2) {$x_1$}; 
    \node[latent, right=of b,draw=orange, label={[label distance=-0.5em, fill=white, draw = orange]30:6}] (c2) {$x_1$}; 
    \node[latent,above=of Middle1,draw =orange, label={[label distance=-0.5em, fill=white, draw = orange]30:2}] (e2) {$\times$}; 
			
		\end{tikzpicture}
		}
		\caption{Donor tree $\mathcal{P}_j$ for variation with $\mathcal{O}_i$.}
	\end{subfigure}
	\newline
	\vspace{1em}
	\begin{subfigure}[t]{0.4\linewidth}
		\centering
		\scalebox{0.5} {
		\begin{tikzpicture}[thick, every node/.style={scale=1.3}, every label/.style=draw]
    \node[latent, draw=none] (a) {$x_1$};%
    \node[latent,right= of a, label={[label distance=-0.5em, fill=white]30:4}] (b) {$x_1$}; %
    \coordinate (Middle1) at ($(a)!0.5!(b)$);
    \node[latent, draw=none, right=of b] (c) {$x_2$}; %
    \node[latent,right=of c, label={[label distance=-0.5em, fill=white]30:7}] (d) {$x_3$}; %
    \coordinate (Middle2) at ($(c)!0.5!(d)$);
    \node[latent, draw=none,above=of Middle1] (e) {$\times$}; %
    \node[latent,above=of Middle2, label={[label distance=-0.5em, fill=white]30:5}] (f) {$-$}; %
    \coordinate (Middle3) at ($(e)!0.5!(f)$);
    \node[latent,above=of Middle3, label={[label distance=-0.5em, fill=white]30:1}] (g) {$+$}; %
     
    % edges
    \draw (g) -- (e);
    \draw (g) -- (f);
    \draw (e) -- (a);
    \draw (e) -- (b);
    \draw (f) -- (c);
    \draw (f) -- (d);
     
    \node[latent, label={[label distance=-0.5em, fill=white]30:3}] (a2) {$x_1$}; 
    \node[latent, right=of b,draw=cyan, label={[label distance=-0.5em, fill=white, draw = cyan]30:6}] (c2) {$x_2$}; 
    \node[latent,above=of Middle1,draw =cyan, label={[label distance=-0.5em, fill=white, draw = cyan]30:2}] (e2) {$-$}; 
			
		\end{tikzpicture}
		}
		\caption{Recipient tree $\mathcal{O}_i$ before variation with $\mathcal{P}_j$.}
	\end{subfigure}
	\hspace{0.5em}
	$\rightarrow$
	\hspace{0.5em}
		\begin{subfigure}[t]{0.4\linewidth}
		\centering
		\scalebox{0.5} {
\begin{tikzpicture}[thick, every node/.style={scale=1.3}, every label/.style=draw]
    \node[latent, draw=none] (a) {$x_1$};%
    \node[latent,right= of a, label={[label distance=-0.5em, fill=white]30:4}] (b) {$x_1$}; %
    \coordinate (Middle1) at ($(a)!0.5!(b)$);
    \node[latent, draw=none, right=of b] (c) {$x_2$}; %
    \node[latent,right=of c, label={[label distance=-0.5em, fill=white]30:7}] (d) {$x_3$}; %
    \coordinate (Middle2) at ($(c)!0.5!(d)$);
    \node[latent, draw=none,above=of Middle1] (e) {$\times$}; %
    \node[latent,above=of Middle2, label={[label distance=-0.5em, fill=white]30:5}] (f) {$-$}; %
    \coordinate (Middle3) at ($(e)!0.5!(f)$);
    \node[latent,above=of Middle3, label={[label distance=-0.5em, fill=white]30:1}] (g) {$+$}; %
     
    % edges
    \draw (g) -- (e);
    \draw (g) -- (f);
    \draw (e) -- (a);
    \draw (e) -- (b);
    \draw (f) -- (c);
    \draw (f) -- (d);
     
    \node[latent, label={[label distance=-0.5em, fill=white]30:3}] (a2) {$x_1$}; 
    \node[latent, right=of b,draw=orange, label={[label distance=-0.5em, fill=white, draw = orange]30:6}] (c2) {$x_1$}; 
    \node[latent,above=of Middle1,draw =orange, label={[label distance=-0.5em, fill=white, draw = orange]30:2}] (e2) {$\times$}; 
			
		\end{tikzpicture}
		}
		\caption{Recipient tree $\mathcal{O}_i$ after variation with $\mathcal{P}_j$.}
	\end{subfigure}
	\hfill
	\caption{GOM step for FOS element $\{2,6\}$ applied to offspring $\mathcal{O}_i$ using donor $\mathcal{P}_j$. }
	\label{fig:GOM}
\end{figure}
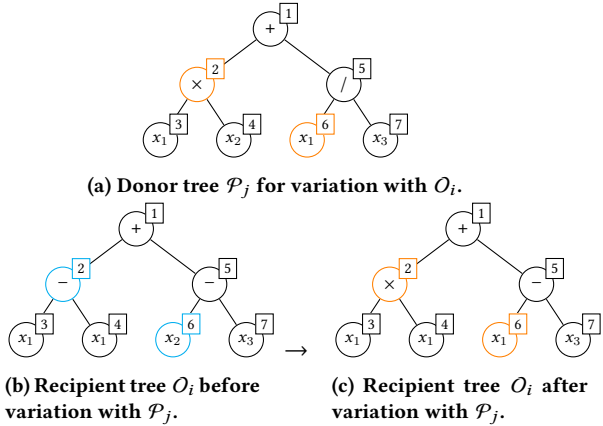

\autoref{fig:GOM} shows an example of a variation step in GP-GOMEA, where we variate $\mathcal{O}_i$ using FOS subset member $\{2,6\}$. Elements at indices $2$ and $6$ in selected individual $\mathcal{O}_i$ are replaced with those of donor tree $\mathcal{P}_j$, where $j$ is randomly selected.

In our approach, we use a multi-tree representation. A multi-tree $T$ consist of multiple trees $t_i$ such that $T = (t_1,\dots,t_n)$. A multi-tree implementation of GP-GOMEA did, however, not yet exist. To support multi-tree in GP-GOMEA, we concatenate the string representations of the $n$ trees. Therefore, the indices count onward from one tree to another tree. For a multi-tree with $2$ trees and a height of $3$, this means that the first tree has a root node with index $1$ and contains indices $1$ through $7$ as in \autoref{fig:GOM}. The second tree has a root node with index $8$ and contains indices $8$ through $14$.

Note that GOM always replaces a gene at index $i$ with a gene in a donor solution at index $i$. GOM will thus not replace nodes from tree $t_i$ of a multi-tree with nodes from $t_j$ from a donor multi-tree where $i\neq j$. We do, however, want to be able to learn interdependencies \emph{across} trees in a multi-tree. This will, for example, enable to learn that index $2$ in the first tree is linked to index $11$ in the second tree, which allows GOM to replace these nodes simultaneously. This is straightforwardly achieved because we concatenate the string representations of the $n$ trees to achieve the string representation in GOMEA. Automatically, therefore, an FOS is learned that can represent interdependencies across trees.

However, we know, by design, that we are modelling multiple trees and that these trees separately will make a substantial contribution to the objective functions. For this reason, it is also of value to learn an FOS (an LT, to be precise) for each GP tree independently, which we propose to do. Moreover, because these GP trees are individual components of the larger multi-tree, we want to be able to exchange \emph{entire trees} as well. For this, when we learn an LT for a single GP tree in the multi-tree representation, we do not discard the FOS element with all variable indices. The final FOS that is used for GOM, is the union of the FOS learned for the entire multi-tree and all of the FOSs learned for each GP tree separately.

\subsection{MO-GP-GOMEA}
Our goal is to search for trees that are both of high-quality and diverse.
Using a multi-tree representation, we can explicitly express these notions over the multi-tree and optimize for them. 

By properly governing that these expressions stay different throughout the search, we could design a multi-modal approach around this representation that searches for $n$ local optima.
However, as stated in the introduction, the expression with the smallest training error is not necessarily the preferred one, due to noise in the data, limited data, missing features, expert knowledge, and/or user needs.
Hence, in a user-centered XAI setting, returning only the expression with the lowest training error, or even multiple expressions with the same optimal training error, is likely not satisfactory.
It would be prudent to accept that our data is not perfect and explicitly also search for other properties of our expressions that represent potentially other things a domain expert could be interested in.
Therefore, we present an objective that stimulates searching for such potentially interesting properties in \autoref{sec:objectives}.

Firstly, however, it is important to realize that finding multiple sets of expressions with different, conflicting properties, requires MO search.
However, only a Single-Objective (SO) version of GP-GOMEA currently exists~\cite{virgolin2021improving}.
We therefore create, for the first time, an MO version of GP-GOMEA, which we
base on the MO version of GOMEA originally introduced for binary variables~\cite{luong2014multi}.

\begin{algorithm}
\caption*{\textbf{MO-GP-GOMEA}(population size $N$, clusters $k$)}
\begin{algorithmic}[1]
\For{$i\in\{0,1,\dots,N-1\}$}
  \State $\mathcal{P}_i\gets$ CreateRandomSolution()
  \State EvaluateFitness($\mathcal{P}_i$)
  \State $\mathcal{A}\gets$ UpdateElitistArchive($\mathcal{P}_i$)
\EndFor
\While{$\neg$TerminationCriteriaSatisfied}
  \State $\{\mathcal{C}_0,\mathcal{C}_1,\dots,\mathcal{C}_{k-1}\}\gets$ ClusterPopulation($\mathcal{P}$)
  \For{$j\in\{0,1,\dots,k-1\}$}
    \State $\mathcal{F}_j\gets$ LearnLinkageModel($\mathcal{C}_j$)
  \EndFor
  \For{$i\in\{0,1,\dots,N-1$\}}
    \State $\mathcal{O}_i\gets$ Clone($\mathcal{P}_i$)
    \State $\mathcal{C}_j\gets$ DetermineCluster($\mathcal{O}_i$, $\{\mathcal{C}_0,\mathcal{C}_1,\dots,\mathcal{C}_{k-1}\}$)
    \If{IsExtremeCluster($\mathcal{C}_j$)}
      \State $\mathcal{O}_i\gets$ SingleObjective-GOM($\mathcal{O}_i$, $\mathcal{C}_j$,$\mathcal{F}_j$,$\mathcal{A}$)
    \Else
      \State $\mathcal{O}_i\gets$ MultiObjective-GOM($\mathcal{O}_i$, $\mathcal{C}_j$,$\mathcal{F}_j$,$\mathcal{A}$)
    \EndIf
    \State $\mathcal{A}\gets$ UpdateElitistArchive($\mathcal{O}_i$)
  \EndFor
  \State $\mathcal{P}\gets\mathcal{O}=\{\mathcal{O}_0,\mathcal{O}_1,\dots,\mathcal{O}_{N-1}\}$
\EndWhile
\end{algorithmic}
\end{algorithm}

Pseudo-code for MO GP-GOMEA is shown above, which at the top level, is essentially analogous to MO-GOMEA~\cite{luong2014multi}. 
Every generation, the population is divided into $k$ clusters (line 6). The clustering approach works on the basis of distance between solutions in normalized objective space, and is described in further detail in the supplementary material.
For every cluster, a separate FOS is learned (line 8).
Restricted mating is applied, meaning that donors for an individual must come from the same cluster as the individual itself.
Then, `extreme' clusters are identified (line 12): these are the clusters that, on average, perform best on a single objective.
In extreme clusters, a single objective is optimized, that is, in GOM only one objective is considered when testing for improvements (line 13).
A change is accepted if the offspring does not have worse fitness than the parent. 
In all other clusters, multi-objective optimization is performed, that is, in GOM both objectives are considered when testing for improvements (line 15).
A change is accepted if the offspring either: 1) dominates the parent, 2) has equal objective values as the parent, 3) is not dominated by any member of the elitist archive and has different objective values than any member in the elitist archive, or 4) has the same objective values as a member in the elitist archive but is different in semantic space.
When variation does not change the solution for one generation or does not improve the solution for $1 +\log_{10} (\text{N})$ generations, we do an additional round of GOM called Forced Improvements (FI).
This means that another round of variation is done with the donor being a random member from the elitist archive, or in the case of extreme clusters, the solution from the archive that has the best fitness for the corresponding single objective.
The acceptance is the same as with regular variation, except, when the offspring has the same objective value(s) as the parent, the change is rejected. In addition, we stop iterating over the FOS once a change is accepted. If it has not changed during FI, we replace the offspring with a random member from the elitist archive, unless it originated from an extreme cluster, in which case we replace it with the solution from the archive that has the best fitness in terms of the corresponding single objective.
After variation, the elitist archive is updated with the offspring (line 16).

\subsection{Objectives}\label{sec:objectives}
We optimize for two objectives, error and diversified error. For error, we use the commonplace Mean Squared Error (MSE). For diversified error, we specify an objective that fits well with the concept of GP-based XAI in practice as outlined in the introduction.
 
\subsubsection{Error objective}
The MSE is arguably the most commonly adopted objective in machine learning, especially for (symbolic) regression. In case of our multi-tree representation, to calculate the error, for each tree $t_i$ of the multi-tree $T$, the MSE is calculated between the targets $y$ associated with data points $X_{j}$ and the predictions of the tree $t_i(X_{j})$.
We calculate the final error $E$ of the multi-tree by summing the MSE of the individual trees, i.e.:

\vspace*{-1mm}
$$
  E = \sum_{i=1}^{n} MSE(t_i, X ,y),
$$
\vspace*{-3mm}
where
$$
  MSE(t_i, X ,y) = \frac{1}{|X|}\sum_{j = 1}^{|X|} (t_i(X_j) - y_j)^2,
$$

\noindent $n$ is the number of trees in a multi-tree, and $|X|$ is the number of records in the data set. 

\subsubsection{Diversified error objective}
As all our experiments will be done with two trees, for now, we introduce the definition for the diversified error objective for a multi-tree with two trees. We come back to this in Section~\ref{sec:discussion}. 
We define the diversified error $D$ as the mean of the minimum squared error of the individual trees, i.e.,:

\vspace*{-1mm}
$$
D = \frac{1}{|X|}\sum_{j = 1}^{|X|} \text{min}((t_1(X_j) - y_j)^2 ,(t_2(X_j) - y_j)^2).
$$

We choose this objective function for three reasons.
Firstly, it enables finding expressions that \emph{together} describe a data set well, i.e., expressions that describe different parts of the data set.
Note that because we also optimize for error objective $E$, we will find expressions that not only predict the data set well together but are also optimized for their individual error.
In addition, describing different parts of a data set with different models can even improve interpretability, because it resonates well with how humans intuitively understand things.
Humans tend to categorize things~\cite{klausmeier1992concept}.
In this way, they can dissect a problem and try to understand parts of it separately.
Secondly, in combination with the error objective, we can find sets of expressions that have a similar error but have a different error distribution over the data points.
Thirdly, an expert may well prefer a model that is very good in some cases, but obviously wrong in others, over a model that is moderately right and wrong in all cases. In practice, this may for instance mean fewer adjustments or further investigations: the cases in which the model is very good do not need to be adjusted. Also, such an objective essentially finds an ensemble of complementary models that would have superior performance upon its use, when it is clear in practice for a particular case which model to follow.
In summary, with this objective, we can find expressions that describe the data better as a set than a single expression would be able to, but at the same time, we also find sets containing expressions with a similar individual error that are different from each other.

Alternatively, one could define a diversity objective that maximizes the distance between the predictions of the trees within a multi-tree.
However, this also stimulates expressions that do not perform well and are just very different. By contrast, using $D$, we do not stimulate diversity merely for the sake of diversity.

Finally, while $D$ is proposed with the goal of offering interesting alternative models, $D$ is a non-smooth non-convex function, which would be unsuitable for most typical machine learning approaches. It is a key strength of GP, being an evolutionary machine learning approach, that it can handle such functions.

\section{Examples on synthetic data}
In this section, we present two examples of the use of our approach on synthetic data.
We generate two simple regression data sets, with Gaussian Noise ($GN$) added. 
We perform a single run with our approach, using multi-tree individuals with $n=2$.
We multi-objectively optimize the error $E$ and the diversified error $D$. 
The allowed symbols are the functions $+, -, \times, \div_{p}$, Ephemeral Random Constants (ERCs), and the input variables that appear in the synthetic data set.
We use a maximum tree height of $3$ ($7$ nodes), and perform $30$ generations with a  population size of $1000$.
We choose this population size because, on the one hand, these are quite simple problems, but on the other hand, we need to search in three `directions'; the two extremes and the trade-off between those two.

\subsection{Example 1: multi-modal data}
We generate a multi-modal data set with one input variable $X$, and target variable $Y$
using the following functions:  

\vspace*{-1mm}
$$
  \begin{array}{lcl}
    f(X) &\!\!\!\! = \!\!\!\!& X^{2} + GN(0,10)\\
    g(X) &\!\!\!\! = \!\!\!\!& 2 \cdot X + GN(0,10)
  \end{array}
$$

We generate 100 data points with $Y=f(X)$ where $X$ is drawn randomly from the interval $X[0,10]$, and generate $40$ data points with $Y=g(X)$ where $X$ is drawn from the same interval.

\begin{figure}[t]
\centering
\includegraphics[width=1\linewidth]{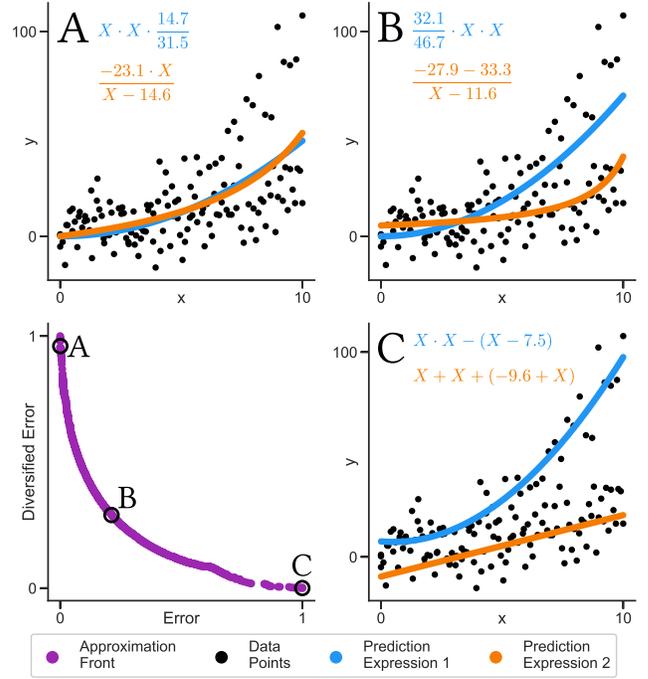}
\vspace*{-3mm}
\caption{The approximation front of our approach on the synthetic multi-modal data set. The predictions of the expressions of multi-trees A, B and C are visualized.}
\label{fig:ens1}
\vspace*{-3mm}
\end{figure}

We show the approximation front of the error and diversified error found by our approach on this data set in \autoref{fig:ens1}. We also show the predicted $Y$ of the two expressions of three multi-tree models along the front, against $X$, as well as the expressions themselves.
As can be observed, our approach can effectively model multi-modal data sets.
In particular, we see that the expressions of multi-tree C closely resemble $f(X)$ and $g(X)$, despite the uneven number of data points per function.
The expressions of multi-tree A fall in between $f(X)$ and $g(X)$, because this minimizes the MSE.
Multi-tree B is somewhere in between A and C.

\subsection{Example 2: hidden variable }
We generate a data set with hidden variable $H$, two input variables $X_1$ and $X_2$, and target variable $Y$, i.e., $H$ itself is not in the data set.
Target variable $Y$ is equal to hidden variable $H$.
Input variables $X_1$ and $X_2$ are flawed `estimations' of $H$. 
More specifically, the data set is generated using the following functions:

\vspace*{-1mm}
$$
  \begin{array}{lcl}
    X_1 &\!\!\!\! = \!\!\!\!& H + GN(0,0.5)\\
    X_2 &\!\!\!\! = \!\!\!\!& H + GN(0,0.5)\\
    Y &\!\!\!\! = \!\!\!\!& H
  \end{array}
$$

\begin{figure}[t]
\centering
\includegraphics[width=1\linewidth]{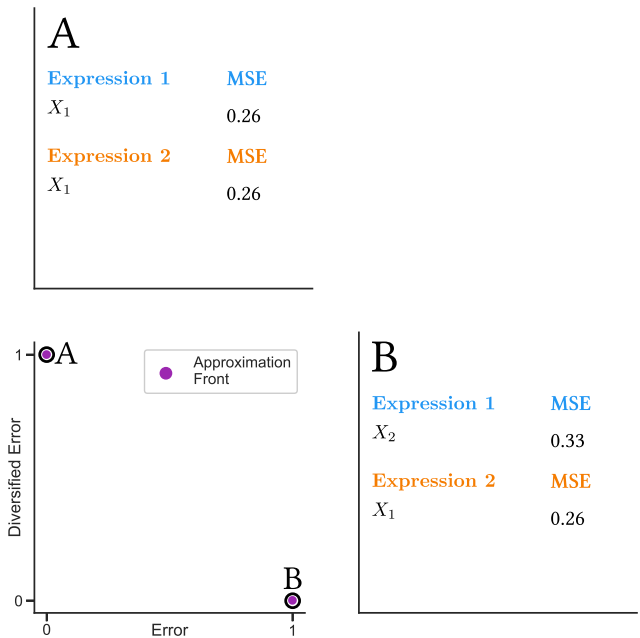}
\vspace*{-6mm}
\caption{The approximation front of our approach on the synthetic hidden variable data set. The two expressions, and the individual MSE of multi-trees A and B are shown. }
\label{fig:appr}
\vspace*{-3mm}
\end{figure}

We generate 100 data points by drawing $H$ randomly from the interval $[0,10]$.
We show the approximation front of the error and diversified error found by our approach on this data set in \autoref{fig:appr}. Furthermore, we show the two expressions, and their individual MSE, of two multi-tree models along the front. 
The parameters are as described above, except here we do not use ERCs. 
By chance, due to the randomness of the Gaussian noise, $X_{1}$ has a slightly smaller error for predicting $Y$.
Therefore, expressions that simply predict $X_1$ have a lower MSE than expressions that predict $X_2$.
However, $X_1$ and $X_2$ are almost equally good. Our approach is able to retrieve multi-tree model B that represents this, with one expression using $X_1$ and the other expression using $X_2$.

The same principle applies to more elaborate data sets as well.
Data sets may have multiple input variables or combinations of input variables that are good predictors of the target variable.
Expressions using different combinations of the variables and functions might have a similar error, but if one of the functions is only slightly better, or if one of the solutions is easier to find when optimizing for MSE only, the focus will be on this solution.
Our approach explicitly stimulates to explore different possibilities and is more likely to find multiple of these solutions.

\section{Evaluation on benchmark data}
In this section, we evaluate our approach on real-world data sets. We first describe our experimental setup before reporting our results.

\subsection{Experimental setup}
We evaluate three approaches: our multi-modal multi-tree MO-GP-GOMEA approach, multi-tree SO optimization with GP-GOMEA, and multi-tree NSGA-II~\cite{deb2002fast}.
We compare with SO optimization to evaluate whether our approach can find solutions at the extremes of the approximation front that are as good as when optimizing for each objective individually.

NSGA-II is a state-of-the-art MO EA, and perhaps the most popular MO EA in literature. Note that the GP version of NSGA-II does not use tree templates but has a variable tree length (with a maximum total size) like in traditional GP. 
We compare with NSGA-II to evaluate whether our approach can find approximation fronts as good as, or better than, a state-of-the-art MO GP algorithm.

We evaluate these approaches on three well-known data sets with a regression task, of which we show the properties in \autoref{tab:datasets}.
The Boston housing data set contains information concerning housing in the area of Boston Mass \cite{harrub78}, where the goal is to predict the median house value in an area.
The concrete compressive strength data set contains information concerning the components and age of concrete \cite{yeh1998modeling}, where the goal is to predict the strength of concrete.
The yacht hydrodynamics data set contains information on the characteristics of yachts \cite{gerritsma1981geometry}, where the goal is to predict the resistance of sailing yachts. 

\begin{table}[t]
\caption{Properties of benchmark data sets.}
\vspace*{-4mm}
\scalebox{0.9}{
\begin{tabular}{ c|c|c }
\toprule
Data set (abbreviation) & Features & Samples\\
\midrule
Boston housing (b) & $13$ & $506$\\
Concrete compressive strength (c) & $9$ & $1030$\\
Yacht hydrodynamics (y) & $6$ & $308$\\

\bottomrule
\end{tabular}}
\label{tab:datasets}
\vspace{-1em}
\end{table}

We initialize the population with multi-tree individuals with $n=2$.
We multi-objectively optimize the two objective functions described in~\autoref{sec:objectives}: $E$ and $D$.
Algorithm parameter and experiment settings are given in \autoref{tab:settings}.
Each run is repeated 30 times with a different random seed. We report the median over these runs and the statistical significance of the Wilcoxon signed-rank test (with $\alpha= 0.05$ and Bonferroni correction).
For NSGA-II we use a crossover proportion of $0.5$, a mutation proportion of $0.5$, a tournament size of $4$, and ramped half-and-half initialization. 

Furthermore, we use a population size of $5000$ for SO optimization,  and a population size of $15000$ in MO optimization. For our approach and NSGA-II, we use a population size that is $3$ times larger than for SO, because MO needs to optimize three `directions', namely the two extremes and the trade-off between those two. For our approach, we use $7$ clusters in MO-GP-GOMEA.

\begin{table}[t]
\caption{Algorithm parameter and experiment settings.}
\label{tab:settings}
\vspace*{-4mm}
\scalebox{0.9}{
\begin{tabular}{c|ccc}
\toprule
Setting                         & \multicolumn{1}{c|}{I}      & \multicolumn{1}{c|}{II}      & III       \\ \midrule
Functions                       & \multicolumn{3}{c}{$+, -, \times, \div_{p}$}                       \\ \midrule
Terminal                        & \multicolumn{3}{c}{variables, ERC}                                 \\ \midrule
Train-test split              & \multicolumn{3}{c}{$75\%$-$25\%$}                                  \\ \midrule
For SO and Ours: tree height            & \multicolumn{1}{c|}{$3$}    & \multicolumn{1}{c|}{$4$}    & $5$     \\ \midrule
For NSGA-II: maximum tree size & \multicolumn{1}{c|}{$7$}    & \multicolumn{1}{c|}{$15$}   & $31$    \\ \midrule
Time (s)                        & \multicolumn{1}{c|}{$1200$} & \multicolumn{1}{c|}{$7200$} & $10800$ \\ \midrule
CPU                             & \multicolumn{3}{c}{AMD EPYC 7282}  \\        
\bottomrule
\end{tabular}}
\end{table}

For both objectives, we report the objective values of the extreme solutions found by the three approaches.
We also report the HyperVolume (HV) of our approach and that of NSGA-II.
HV is a measure that indicates the volume covered by the approximation front with respect to a reference point.
We compute the HV by first combining all approximation fronts found by the different approaches over all runs, and then extracting the non-dominated solutions, i.e., we take the front of fronts.
Then, we take the minimum and maximum values in each objective from this front, and we use them to normalize all approximation fronts.
Finally, we get the HV by computing the surface area covered by the front with respect to reference point $[1.1,1.1]$. Note that this can mean that HV values of 0 can be reported, indicating that the median run did not find any multi-tree solution near the best found solutions.

\newcommand{\bd}{\begingroup{\color{blue}\blacktriangledown}\endgroup}
\newcommand{\cd}{\begingroup{\color{cyan}\blacktriangledown}\endgroup}
\newcommand{\md}{\begingroup{\color{magenta}\blacktriangledown}\endgroup}

\newcommand{\bu}{\begingroup{\color{blue}\blacktriangle}\endgroup}
\newcommand{\cu}{\begingroup{\color{cyan}\blacktriangle}\endgroup}

 \subsection{Results}
We report the results of the experiment setting II in Tables \ref{tab:hv} and \ref{tab:extreme}. 
Table \ref{tab:hv} shows that generally our approach has a significantly bigger HV than NSGA-II.  Table \ref{tab:extreme} shows that the best performing solution of our approach generally has no significant difference in diversified error $D$ compared with SO, and is significantly better than NSGA-II. Similar findings can be observed in Table  \ref{tab:extreme} regarding error $E$. The supplementary material includes the results of experiment I and III. These results generally show the same pattern as in experiment II. For the error objective, however, our approach is significantly better than SO in experiment I, whereas SO is significantly better in many cases in experiment III. The median values however, are similar. From the above, we conclude that we can effectively find a high-quality approximation front that includes the extreme solutions.

\begin{table}[t]
\caption{Median HV results for experiment setting II. A triangle symbol next to the reported median value indicates significant superiority (better (=bigger) HV) to the approach with the name in the same color as the triangle.}
\vspace*{-3mm}
\scalebox{0.9}{
\begin{tabular}{ll|rl|rl}
\toprule
Data set & Split &  \textcolor{cyan}{Ours} &  &  \textcolor{blue}{NSGA-II} &  \\
\midrule
      b & Train &       0.61 &        $\bu$ &       0.33 &              \\
      b &  Test &       0.00 &              &       0.00 &              \\
      c & Train &       0.57 &        $\bu$ &       0.00 &              \\
      c &  Test &       0.32 &        $\bu$ &       0.00 &              \\
      y & Train &       0.11 &        $\bu$ &       0.00 &              \\
      y &  Test &       0.00 &        $\bu$ &       0.00 &              \\
\bottomrule
\end{tabular}}
\label{tab:hv}
\end{table}

\begin{table}[t]
\caption{Median best diversified error $D$ and median best error $E$ for experiment setting II. A down-pointing triangle next to the reported median value indicates significant superiority (better (=smaller) objective value) to the approach with the name in the same color as the down-pointing triangle.}
\vspace*{-3mm}
\scalebox{0.9}{
\begin{tabular}{ll|rll|rl|rll}
\toprule
Data set & Split &  \textcolor{cyan}{Ours} &  &  &  \textcolor{blue}{NSGA-II} &    &  \textcolor{magenta}{SO} &  &  \\
\midrule
\multicolumn{10}{c}{$D$} \\
\midrule
      b & Train &             6.97 &              $\bd$ &                    &             9.58 &                                        &             6.77 &               &              $\bd$ \\
      b &  Test &            13.29 &                    &                    &            13.86 &                                        &            13.24 &                    &                    \\
      c & Train &            31.98 &              $\bd$ &              $\md$ &            46.67 &                                        &            34.30 &                    &              $\bd$ \\
      c &  Test &            32.71 &              $\bd$ &                    &            46.48 &                                        &            35.79 &                    &              $\bd$ \\
      y & Train &             0.96 &              $\bd$ &                    &             3.36 &                                        &             1.10 &                    &              $\bd$ \\
      y &  Test &             1.44 &              $\bd$ &                    &             2.94 &                                        &             1.66 &                    &              $\bd$ \\
\midrule
\multicolumn{10}{c}{$E$} \\
\midrule
      b & Train &       39.17 &         $\bd$ &               &       48.96 &                              &       39.33 &               &         $\bd$ \\
      b &  Test &       51.13 &         $\bd$ &               &       61.78 &                              &       51.38 &               &         $\bd$ \\
      c & Train &      191.79 &         $\bd$ &               &      279.83 &                              &      194.37 &               &         $\bd$ \\
      c &  Test &      211.41 &         $\bd$ &               &      274.80 &                              &      195.65 &         $\cd$ &         $\bd$ \\
      y & Train &        7.17 &         $\bd$ &               &       24.39 &                              &        7.23 &               &         $\bd$ \\
      y &  Test &        8.87 &         $\bd$ &               &       19.74 &                              &        8.23 &               &         $\bd$ \\
\bottomrule

\end{tabular}}
\label{tab:extreme}
\vspace{-2em}
\end{table}

\section{Examples on real-world data}
In this section, we take a closer look at some results of our approach, using the yacht and the Boston housing data sets as examples.
These examples illustrate how our approach could be useful to users.
For both examples, we perform a single run and describe the expressions found by our approach.
We use the settings as described in our experimental setup in the previous section, with tree height $3$.

\subsection{Example 1: Yacht data set}
\begin{figure}[t]
\centering
\includegraphics[width=1\linewidth]{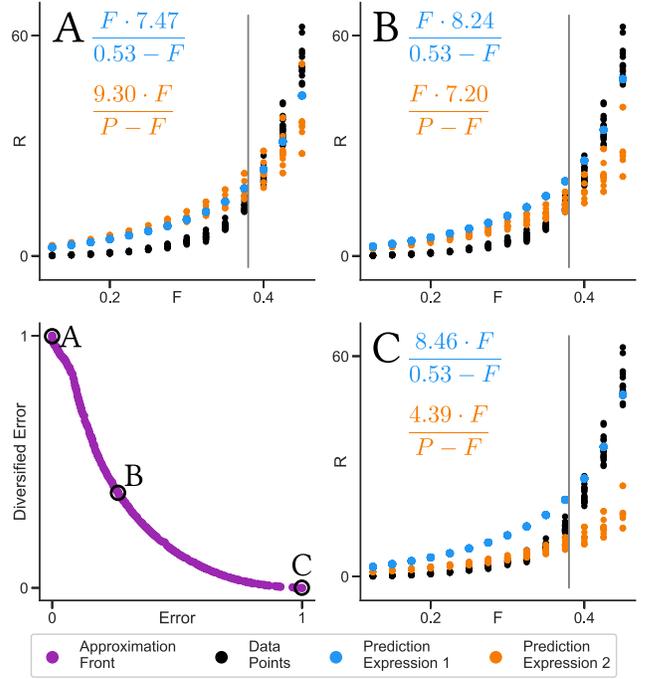}
\vspace*{-6mm}
\caption{The approximation front of our approach on the yacht data set. The predictions $\hat{R}$ of the expressions of multi-trees A, B, and C are visualized against $F$.}
\label{fig:appr2}
\vspace*{-3mm}
\end{figure}

In \autoref{fig:appr2}, we show the approximation front of the error and diversified error found by our approach on the yacht data set. 
We also show the predicted resistance $\hat{R}$ of the two expressions of three multi-tree models along the front, against the Froude number $F$, as well as the expressions themselves.
The Froude number is a ratio between the speed of the yacht and the gravity, which is known to influence the (wave) resistance of a yacht.
$P$ is the prismatic coefficient, which is a measure of how quick the breadth of the yacht
changes or in other words a measure for the fullness of the ends
of the yacht.

We see that the expressions of multi-tree C, which is an extreme solution, having the best diversified error $D$, describe the data well together: expression 1 describes the data well for $F > 0.38$, while expression 2 describes the data well for $F < 0.38$.
This corresponds well to findings in literature: according to the theory of Chapman, as well as experimental results, the resistance increases drastically when the Froude number exceeds approximately $0.38$~\cite{tuck1987wave, chapman1972hydrodynamic}.

The expressions of multi-tree A, on the other hand, which is an extreme solution, having the best error $E$, both try to model the whole data set, which corresponds to having the lowest MSE.
However, neither accurately describes the data for either $F < 0.38$ or $F > 0.38$.
The predictions of the expressions of multi-tree B are visually somewhere in between the predictions of A and C.

%This example illustrates how multiple expressions can be useful when describing a problem.

\subsection{Example 2: Boston housing data set}
\begin{figure}[t]
    \centering
    \includegraphics[width=1\linewidth]{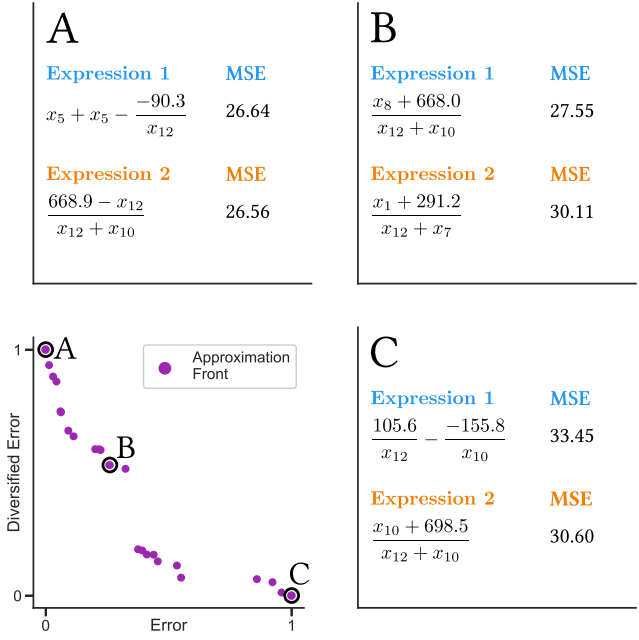}
    \vspace*{-6mm}
    \caption{The approximation front of our approach on the Boston housing data set. The two expressions, and the individual MSE of multi-trees A, B, and C are shown. $x_1$ is the proportion of land with big lots,  $x_5$ is the average number of rooms per residence, $x_7$ is the weighted distance to five employment centers, $x_8$ is the accessibility to radial highways, $x_{10}$ is the ratio of pupils to teachers of a town, and $x_{12}$ is the percentage of adults without high school education.}
    \label{fig:appr_boston}
    \vspace*{-6mm}
\end{figure}

In \autoref{fig:appr_boston}, we show the approximation front of the error and diversified error found by our approach on the Boston housing data set. Furthermore, we show the two expressions, and their individual MSE, of three multi-tree models along the front. 

Expression 2 of multi-tree A has the lowest error.
However, depending on the task, this expression might not be the expression that best accommodates the needs of the user.
Assume the user is a real estate investor that is interested in the factors that influence the house price.
Now suppose that the user knows that a new highway will be built in the near future.
The user wants to take this into account when buying new houses and wants to predict how the highway might affect the value of houses in that area, to predict how much profit can be made.
Therefore, the user employs expression 1 of multi-tree B, which uses $x_8$, the index of the accessibility to radial highways, assuming the new highway only effects $x_8$ and not the other variables, instead of expression 2 of multi-tree A that does not use $x_8$ and is only slightly different in MSE. This shows how it can be useful to search for multiple expressions that are different but have a similar MSE. If the user had used SO optimization to generate expressions, the user would not have had this choice, because only one model would have been presented.

\section{Discussion}\label{sec:discussion}
We evaluated our approach with multi-trees where $n=2$.
To use our approach with multi-trees where $n>2$, we need to generalize the diversified error objective function $D$. In principle, this can be done by simply taking the minimum error over all trees. For the extreme solution that optimizes this generalization, individual trees will still represent different parts of the data more closely. Toward the other extreme, the sum of MSE values, however, this generalization does not necessarily have the same effect as for $n=2$. To realize this also for $n>2$, we specify \emph{two} diversified error functions, $D_1$ and $D_2$.
$D_1$ is the aforementioned generalization, and $D_2$ is the average \emph{pairwise} mean of the minimum squared error of the trees:

\vspace*{-3mm}
$$
  \begin{array}{lcl}
  D_1 &\!\!\!\! = \!\!\!\!&\displaystyle\frac{1}{|X|}\sum_{j = 1}^{|X|} \text{min}((t_1(X_j) - y_j)^2 ,\dots , (t_n(X_j) - y_j)^2)\\[-1mm]
  D_2 &\!\!\!\! = \!\!\!\!&\displaystyle\frac{2}{n\cdot (n-1)}\sum_{i = 1}^{n-1} \sum_{l = i+1 }^{n} D_p (t_i,t_l), \\[-3mm]
  \end{array}
$$

where

\vspace*{-3mm}
$$
    D_p (t_i,t_l) =\frac{1}{|X|}\sum_{j = 1}^{|X|} \text{min}((t_i(X_j) - y_j)^2 , (t_l(X_j) - y_j)^2).
$$

\begin{figure}[ht]
\vspace*{-3mm}
\includegraphics[width=0.5\linewidth]{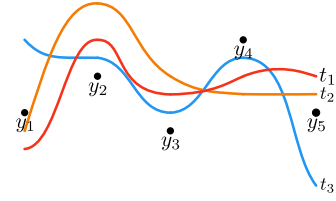}
\vspace*{-3mm}
\caption{A multi-model with $n=3$ and target values $y_1,\dots,y_5$.}
\label{fig:diversity_error_example}
\vspace*{-3mm}
\end{figure}

Note that for $n = 2$ the objective functions are equal, i.e., $D = D_1 = D_2$.
$D_1$ stimulates sets of expressions that describe a data set well together.
$D_2$ stimulates sets of expressions that have similar error but have a different error distribution over the data points.
We explain why $D_2$ stimulates a different kind of diversity with an example in \autoref{fig:diversity_error_example}. In this example, either $t_2$ or $t_3$ is closer to every target value $y_1,\dots,y_5$ than $t_1$. Therefore, $t_1$ does not decrease $D_1$, even though it adds diversity with respect to $t_2$ and $t_3$. $D_2$, in contrast, takes into account the diversity that $t_1$ adds.

When $n>2$, there are a few options regarding how to use $D_1$ and $D_2$ alongside $E$.
One could choose to 1) use either $D_1$ or $D_2$, depending on their needs, 2) perform multiple runs, some of which optimize $D_1$, and some of which optimize $D_2$, or 3) perform a run which optimizes both $D_1$ and $D_2$ along $E$ with MO optimization with three objectives. This may however make final model selection more complicated.

Finally, an important next step is to evaluate our approach in a real-world setting where users are provided with multiple expressions to choose from. For example, it could be researched how people interact with the solutions that our approach finds, and how to present multiple solutions to a user. Another aspect to study is how to aggregate and/or combine the results of our approach with  cross-validation, given that you would get multiple fronts of models. Finally, the interleaved multi-start scheme~\cite{dushatskiy2021parameterless} could be implemented such that users do not have to choose a population size when using our approach. 

\section{Conclusion}
In this work, we have presented a novel multi-modal multi-tree MO GP approach that extends a modern model-based GP algorithm known as GP-GOMEA.
We presented experimental evidence on synthetic and real-world data that showed that our approach can generate multiple diverse high-quality expressions that include expressions that excel in different notions of quality.
Our approach could be a promising approach to allow users to inspect different models, which could lead to novel insights into the data and the process underlying the data.
Providing the user with options in this manner, possibly in combination with additional expert knowledge, could help support them in choosing a good model for the task at hand in a powerful and sensible way.

\begin{acks}
This research was funded by the European Commission within the HORIZON Programme (TRUST AI Project, Contract No.: 952060).
\end{acks}

\end{document}

\title{Multi-modal multi-objective model-based genetic programming to find multiple diverse high-quality models --- Supplementary materials}

\author{E.M.C. Sijben}
\affiliation{%
 \institution{Centrum Wiskunde \& Informatica}
 \city{Amsterdam}
 \country{the Netherlands}
}
\email{evi.sijben@cwi.nl}

\author{T. Alderliesten }
\affiliation{%
 \institution{Leiden University Medical Center}
 \city{Leiden}
 \country{the Netherlands}
}
\email{t.alderliesten@lumc.nl}

\author{P.A.N. Bosman }
\affiliation{%
 \institution{Centrum Wiskunde \& Informatica}
 \city{Amsterdam}
 \country{the Netherlands}
}
\email{peter.bosman@cwi.nl}

\maketitle

\section{Clustering in MO-GOMEA }
\begin{figure*}[h]
\includegraphics[width=0.8\textwidth]{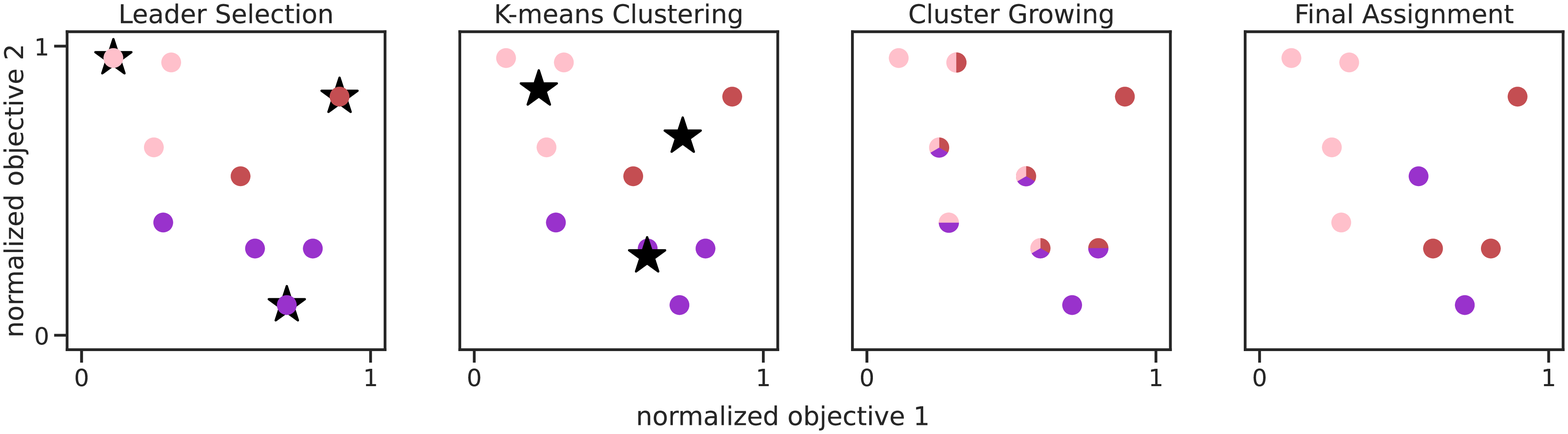}
\caption{Example of how BKLM operates. The points represent the normalized objective values of the individuals. In ``Leader Selection", stars denote the positions of the leaders. In ``K-means Clustering", stars denote the position of the cluster centers. The color of a point indicates the cluster that the individual belongs to. If a point has multiple colors this means that the point belongs to multiple clusters. We illustrate a very small population size to keep the figure readable. }

\label{fig:clustering}
\end{figure*}

We use the same clustering method in our approach as in MO-GOMEA~\cite{luong2014multi}, which is Balanced K-Leader-Means (BKLM) clustering.
First, $k$ leaders are selected.
The first leader is selected by taking the individual with the minimum value for a randomly chosen objective.
Next, the distances $\mathcal{D}=(\mathcal{D}_1,\dots,\mathcal{D}_N)$ between each individual and the first leader are computed, with $N$ being the population size.
The individual with the maximum distance is selected as the next leader.
Then, the distances $\mathcal{D}$ are updated by taking for each individual $\mathcal{P}_i$ the minimum of $\mathcal{D}_i$ and the distance between $\mathcal{P}_i$ and the new leader.
This is repeated until $k$ leaders are found.
These $k$ leaders are used as initial cluster centers to perform $k$-means clustering.
After $k$-means clustering has converged, each cluster is assigned the $\frac{2\cdot N}{k}$ solutions that are closest to its center. For variation, however, it needs to be known for each individual which FOS to use, which donors to choose from, and whether to use SO GOM or MO GOM.
Therefore, each individual needs to be assigned to precisely one cluster.
To do this, individuals that are not yet assigned to a cluster, are assigned to the cluster with the closest center, and individuals that are assigned to multiple clusters, are assigned to one of these clusters at random. \autoref{fig:clustering} illustrates this process.

\newpage

\section{Experimental results}

In the main text, we describe three different settings for our experiments: I, II, and III. We show results of these experiments in Tables \ref{tab:HV}, \ref{tab:D}, and \ref{tab:E}. 
\begin{table}[h]
\caption{Median HV results for experiment settings I, II, and III. A triangle symbol next to the reported median value indicates significant superiority (better (=bigger) HV) to the approach with the name in the same color as the triangle.}
\begin{tabular}{ll|rl|rl}
\toprule
Data set & Split &  \textcolor{cyan}{Ours} &  &  \textcolor{blue}{NSGA-II} &  \\
\midrule
 \multicolumn{6}{c}{Setting I} \\     
\midrule
      b & Train &       0.17 &        $\bu$ &       0.00 &              \\
      b &  Test &       0.05 &        $\bu$ &       0.01 &              \\
      c & Train &       0.03 &        $\bu$ &       0.00 &              \\
      c &  Test &       0.35 &        $\bu$ &       0.12 &              \\
      y & Train &       0.00 &              &       0.00 &              \\
      y &  Test &       0.00 &              &       0.00 &              \\
\midrule
 \multicolumn{6}{c}{Setting II} \\     
\midrule
      b & Train &       0.61 &        $\bu$ &       0.33 &              \\
      b &  Test &       0.00 &              &       0.00 &              \\
      c & Train &       0.57 &        $\bu$ &       0.00 &              \\
      c &  Test &       0.32 &        $\bu$ &       0.00 &              \\
      y & Train &       0.11 &        $\bu$ &       0.00 &              \\
      y &  Test &       0.00 &        $\bu$ &       0.00 &              \\
      \midrule
 \multicolumn{6}{c}{Setting III} \\ 
 \midrule
      b & Train &       0.57 &        $\bu$ &       0.09 &              \\
      b &  Test &       0.00 &              &       0.00 &              \\
      c & Train &       0.57 &        $\bu$ &       0.00 &              \\
      c &  Test &       0.33 &        $\bu$ &       0.00 &              \\
      y & Train &       0.45 &        $\bu$ &       0.00 &              \\
      y &  Test &       0.00 &              &       0.00 &              \\
\bottomrule
\end{tabular}
\label{tab:HV}
\end{table}
\newpage

\begin{table}[h]
\caption{Median best diversified error $D$ for experiment settings I,II and III. A down-pointing triangle next to the reported median value indicates significant superiority (better (=smaller) objective value) to the approach with the name in the same color as the down-pointing triangle.}
\begin{tabular}{ll|rll|rll|rll}
\toprule
Data set & Split &  \textcolor{cyan}{Ours} &  &  &  \textcolor{blue}{NSGA-II} &  &  &  \textcolor{magenta}{SO} &  &  \\
\midrule
 \multicolumn{10}{c}{Setting I} \\     
\midrule
      b & Train &            10.40 &              $\bd$ &                    &            14.02 &                    &                    &            10.72 &                    &              $\bd$ \\
      b & Test &            15.17 &              $\bd$ &                    &            18.31 &                    &                    &            15.12 &                    &              $\bd$ \\
      c & Train &            63.04 &              $\bd$ &              $\md$ &            83.32 &                    &                    &            68.02 &                    &              $\bd$ \\
      c &  Test &            61.67 &              $\bd$ &                    &            81.73 &                    &                    &            65.59 &                    &              $\bd$ \\
      y & Train &             6.65 &              $\bd$ &              $\md$ &            14.46 &                    &                    &            13.79 &                    &              $\bd$ \\
      y &  Test &             7.09 &              $\bd$ &              $\md$ &            11.22 &                    &                    &            13.27 &                    &                    \\

\midrule
 \multicolumn{10}{c}{Setting II} \\     
\midrule
      b & Train &             6.97 &              $\bd$ &                    &             9.58 &                    &                    &             6.77 &                    &              $\bd$ \\
      b &  Test &            13.29 &                    &                    &            13.86 &                    &                    &            13.24 &                    &                    \\
      c & Train &            31.98 &              $\bd$ &              $\md$ &            46.67 &                    &                    &            34.30 &                    &              $\bd$ \\
      c &  Test &            32.71 &              $\bd$ &                    &            46.48 &                    &                    &            35.79 &                    &              $\bd$ \\
      y & Train &             0.96 &              $\bd$ &                    &             3.36 &                    &                    &             1.10 &                    &              $\bd$ \\
      y &  Test &             1.44 &              $\bd$ &                    &             2.94 &                    &                    &             1.66 &                    &              $\bd$ \\
\midrule
\multicolumn{10}{c}{Setting III} \\ 
\midrule
      b & Train &             5.09 &              $\bd$ &                    &             6.95 &                    &                    &             5.02 &                    &              $\bd$ \\
      b &  Test &            11.83 &                    &                    &            13.52 &                    &                    &            12.81 &                    &                    \\
      c & Train &            20.70 &              $\bd$ &                    &            30.31 &                    &                    &            21.19 &                    &              $\bd$ \\
      c &  Test &            20.24 &              $\bd$ &                    &            32.59 &                    &                    &            21.28 &                    &              $\bd$ \\
      y & Train &             0.63 &              $\bd$ &                    &             1.21 &                    &                    &             0.63 &                    &              $\bd$ \\
      y &  Test &             1.32 &              $\bd$ &                    &             1.47 &                    &                    &             1.22 &                    &              $\bd$ \\
\bottomrule
\end{tabular}
\label{tab:D}
\end{table}

\begin{table}[t]
\caption{Median best error $E$ for experiment settings I,II and III. A down-pointing triangle next to the reported median value indicates significant superiority (better (=smaller) objective value) to the approach with the name in the same color as the down-pointing triangle.}

\begin{tabular}{ll|rll|rll|rll}
\toprule
Data set & Split &  \textcolor{cyan}{Ours} &  &  &  \textcolor{blue}{NSGA-II} &  &  &  \textcolor{magenta}{SO} &  &  \\
\midrule
\multicolumn{10}{c}{Setting I} \\     
\midrule
      b & Train &       53.26 &         $\bd$ &         $\md$ &       81.54 &               &               &       55.72 &               &         $\bd$ \\
      b &  Test &       65.37 &         $\bd$ &         $\md$ &      105.70 &               &               &       69.37 &               &         $\bd$ \\
      c & Train &      376.19 &         $\bd$ &         $\md$ &      464.92 &               &               &      385.03 &               &         $\bd$ \\
      c &  Test &      399.43 &         $\bd$ &               &      448.56 &               &               &      410.08 &               &         $\bd$ \\
      y & Train &       65.89 &         $\bd$ &         $\md$ &      111.09 &               &               &       89.20 &               &         $\bd$ \\
      y &  Test &       65.78 &         $\bd$ &         $\md$ &      103.78 &               &               &       83.93 &               &         $\bd$ \\
\midrule
\multicolumn{10}{c}{Setting II} \\     
\midrule
      b & Train &       39.17 &         $\bd$ &               &       48.96 &               &               &       39.33 &               &         $\bd$ \\
      b &  Test &       51.13 &         $\bd$ &               &       61.78 &               &               &       51.38 &               &         $\bd$ \\
      c & Train &      191.79 &         $\bd$ &               &      279.83 &               &               &      194.37 &               &         $\bd$ \\
      c &  Test &      211.41 &         $\bd$ &               &      274.80 &               &               &      195.65 &         $\cd$ &         $\bd$ \\
      y & Train &        7.17 &         $\bd$ &               &       24.39 &               &               &        7.23 &               &         $\bd$ \\
      y &  Test &        8.87 &         $\bd$ &               &       19.74 &               &               &        8.23 &               &         $\bd$ \\
\midrule
\multicolumn{10}{c}{Setting III} \\ 
\midrule
      b & Train &       32.32 &         $\bd$ &               &       39.54 &               &               &       32.05 &               &         $\bd$ \\
      b &  Test &       47.48 &         $\bd$ &               &       54.19 &               &               &       45.99 &         $\cd$ &         $\bd$ \\
      c & Train &      121.25 &         $\bd$ &               &      182.77 &               &               &      119.15 &               &         $\bd$ \\
      c &  Test &      122.05 &         $\bd$ &               &      185.57 &               &               &      122.42 &               &         $\bd$ \\
      y & Train &        4.99 &         $\bd$ &               &        8.65 &               &               &        4.20 &         $\cd$ &         $\bd$ \\
      y &  Test &        6.51 &         $\bd$ &               &        8.21 &               &               &        5.73 &         $\cd$ &         $\bd$ \\

\bottomrule
\end{tabular}
\label{tab:E}
\end{table}